\documentclass{INTERSPEECH2023}
\usepackage{cite}
\DeclareMathAlphabet\mathbfcal{OMS}{cmsy}{b}{n}
\usepackage{multirow}
\usepackage{multicol}

\def\mat#1{\mathbf{#1}}
\def\ten#1{\mathbfcal{#1}}

\interspeechcameraready

\title{Quantization-Aware and Tensor-Compressed Training of Transformers for Natural Language Understanding}
\name{Zi Yang$^1$, Samridhi Choudhary$^2$, Siegfried Kunzmann$^2$, Zheng Zhang$^1$}
\address{
 $^1$ Department of Electrical \& Computer Engineering, University of California, Santa Barbara, CA\\
  $^2$ Amazon Alexa AI }
\email{ziy@ucsb.edu, samridhc@amazon.com, kunzman@amazon.com, zhengzhang@ece.ucsb.edu}

\begin{document}

\maketitle
 
\begin{abstract}
Fine-tuned transformer models have shown superior performances in many natural language tasks. However, the large model size prohibits deploying high-performance transformer models on resource-constrained devices. This paper proposes a quantization-aware tensor-compressed training approach to reduce the model size, arithmetic operations, and ultimately runtime latency of  transformer-based models. We compress the embedding and linear layers of transformers into small low-rank tensor cores, which significantly reduces model parameters. A quantization-aware training with learnable scale factors is used to further obtain low-precision representations of the tensor-compressed models. The developed approach can be used for both end-to-end training and distillation-based training. To improve the convergence, a layer-by-layer distillation is applied to distill a quantized and tensor-compressed student model from a pre-trained transformer. The performance is demonstrated in two natural language understanding tasks, showing up to $63\times$ compression ratio, little accuracy loss and remarkable inference and training speedup. 

\end{abstract}

\section{Introduction}

Transformer models~\cite{vaswani2017attention} have been widely used for natural language understanding (NLU)~\cite{devlin2019bert,antoun2020arabert,radfar2020end} and automatic speech recognition (ASR)~\cite{moritz2020streaming,zhang2021usefulness,kimsqueezeformer}. Typically, larger pre-trained transformer models perform better on downstream tasks~\cite{devlin2019bert, liu2019roberta, yang2019xlnet,baevski2020wav2vec}. However, these large-size models cannot be deployed directly on edge devices due to the limited computing, memory, and energy resources as well as low latency requirement. As a result, model compression has become an indispensable step to enable efficient deployment of large NLU and ASR models on resource-constraint hardware platforms~\cite{mysore-sathyendra-etal-2020-extreme, saade2018spoken, sun2020mobilebert}.  Existing works have studied NLU and ASR model compression via knowledge distillation \cite{sanh2019distilbert,aguilar2020knowledge,jiao2019tinybert,hou2020dynabert},  quantization\cite{zhang2020ternarybert,bai2020binarybert,shen2020q} and low-rank matrix factorization~\cite{Saghir2021,mao-etal-2020-ladabert}. Among these approaches, low-rank matrix compression normally achieves much higher compression ratios.

Meanwhile, studies in the applied math community have shown that tensor decomposition~\cite{tensor:suvey} often achieves a much higher compression ratio than matrix compression approaches. As a high-dimensional generalization of matrix decompositions, low-rank tensor decomposition has achieved state-of-the-art results in neural network compression~\cite{novikov2015,hawkins2022towards,hawkins2021bayesian,tjandra2017compressing}, including both post-training compression and end-to-end compressed training. Recently, tensor decomposition has also been employed to compress transformer models used in natural language modeling~\cite{ma2019tensorized}. Since many edge devices (e.g., embedded CPU, embedded GPU and FPGA) support low-precision computation, it is natural to ask if low-precision tensor compression can be used to achieve further cost reduction on edge devices. A previous study~\cite{zhang2021fpga} investigated low-precision training of tensor-compressed models, but it shows that directly applying existing low-precision training in the tensor-compressed setting can cause a remarkable accuracy drop even on a simple two-layer perceptron network.

In this work, we present a quantization-aware and tensor-compressed training approach for transformers. We first use low-rank tensor train (TT) and tensor-train matrix (TTM) formats to represent the embedding tables and linear layers respectively, which achieve significant parameter reduction. 
To further reduce memory and computing costs, we 
apply quantization-aware training with learnable scale factors, which enforces the low-rank tensor factors of transformer models into low precision. Our work uses 2-, 4-, or 8-bit fixed-point uniform quantization. 
The proposed quantization-aware and tensor-compressed training can be used for both end-to-end training and post-training compression. In order to leverage the information of pre-trained transformer models to save the training cost, we further employ layer-by-layer distillation \cite{aguilar2020knowledge} to match the internal outputs and attention probabilities of the original model and our low-precision tensor-compressed model to maintain the generalization capability. This layer-by-layer distillation can avoid the divergence issue of distilling all layers in a tensor-compressed format.  
We demonstrate the quantization-aware and tensor-compressed training approach on NLU tasks, ATIS dataset \cite{hemphill1990atis} and GLUE benchmark \cite{wang2018glue}. We perform end-to-end training on the ATIS dataset and compress the BERT-base via layer-by-layer distillation on the GLUE benchmark. In both tasks, our approach reaches ultra-low model size with little performance degradation. 



\section{Methodology}
\subsection{Tensor-Compressed Transformer Training}
\label{subsec:tensor}

A typical transformer model \cite{vaswani2017attention} consists of an embedding table and a set of encoder blocks, where each encoder has one self-attention layer and one feed-forward layer. All self-attention and feed-forward layers are composed of linear layers. The embedding table can be regarded as a special type of linear layer. Tensor-compressed transformer compresses the weight matrices of the linear layers into small tensor cores. We directly train the small tensor cores rather than larger weight matrices.


Consider the linear layer $\mat{y}=\mat{W}\mat{x}+\mat{b}$, where $\mat{x} \in \mathbb{R}^N$ is the input, $\mat{W}\in \mathbb{R}^{M\times N}$ is the weight matrix, and $\mat{b}\in \mathbb{R}^M$ is the bias vector. The weight $\mat{W}$ is reshaped into a tensor $\ten{W}\in \mathbb{R}^{m_1\times \cdots \times m_d \times n_1\times \cdots n_d}$, where $\Pi_{i=1}^d m_i = M$ and $\Pi_{i=1}^d n_i = N$. Then we employ either the tensor-train (TT) format or tensor-train matrix (TTM) format to reduce the number of model parameters. 

 {\bf TT Compression.} The TT format represents tensor $\ten{W}$ as a set of small-size tensor cores $\ten{G}_1,\ldots,\ten{G}_{2d}$, where $\ten{G}_i \in \mathbb{R}^{r_{i-1}\times m_i \times r_{i}}$ for $1\le i\le d$ and $\ten{G}_i \in \mathbb{R}^{r_{i-1}\times n_{i-d} \times r_{i}}$ for $d+1\le i\le 2d$. The tensor $\ten{W}$ and the tensor cores $\{\ten{G}_i\}_{i=1}^{2d}$ satisfy the following equation 
\[
    \ten{W}(i_1,\ldots,i_d,j_1,\ldots,j_d) = \mat{G}_1^{i_1}\cdots\mat{G}_d^{i_d} \mat{G}_{d+1}^{j_1} \cdots\mat{G}_{2d}^{j_d},
\]
where $\mat{G}_k^{i_k}:=\ten{G}_k(:,i_k,:) \in \mathbb{R}^{r_{k-1}\times r_k}$ and $\mat{G}_{k+d}^{j_k}:=\ten{G}_{k+d}(:,j_k,:) \in \mathbb{R}^{r_{k-1+d}\times r_{k+d}}$.
The tuple $(r_0,r_1,\ldots,r_{2d})$ is called the TT rank, with $r_0=r_{2d}=1$. 
The TT-compressed linear layer stores the small tensor cores $\{\ten{G}_i\}_{i=1}^{2d}$ rather than the large matrix $\mat{W}$. After compression, the number of model parameters is reduced to $\sum_{i=1}^d (r_{i-1}m_i r_i + r_{i-1+d}n_i r_{i+d})$ from $MN = m_1\cdots m_d n_1\cdots n_d$. For fixed ranks, the reduction is roughly $O(m_1\cdots m_d n_1\cdots n_d)\to O(\sum_{i=1}^d(m_i+n_i)) $. The compression ratio is determined by the TT rank. For the convenience of discussions and experiments, we fix the TT rank before training, but the TT ranks can also be determined automatically in the training process~\cite{hawkins2022towards}.  The matrix-vector multiplication in TT format can be done efficiently with fewer arithmetic operations than standard matrix-vector products \cite{liu2022tt}.

 {\bf TTM Compression.} The TTM decomposition represents tensor $\ten{W}$ as $d$ tensor cores $\{\ten{F}_i \in \mathbb{R}^{p_{i-1}\times m_i \times n_i \times p_i}\}_{i=1}^d$. The tensor cores satisfy 
\[
    \ten{W}(i_1,\ldots,i_d,j_1,\ldots,j_d) = \mat{F}_1^{i_1,j_1}\cdots\mat{F}_d^{i_d,j_d},
\]
where $\mat{F}_k^{i_k,j_k}:=\ten{F}_k(:,i_k,j_k,:) \in \mathbb{R}^{p_{i-1}\times p_i}$.


The matrix-vector product using TT format is faster TTM format since the contraction order for TT format is optimized as in \cite{liu2022tt}. The TTM compression is more suitable for weight matrices with unbalanced rows and columns. In our tensor-compressed transformer, all linear layers in encoder blocks are trained in the TT format for efficient computation, and the embedding table is trained in the TTM format since the number of rows is much larger than the number of columns.

Assume that the weights and embedding tables $\{\mat{W}_j\}_{j=1}^M$ of a transformer are represented with a set of small tensor cores $\{\ten{G}_i\}_{i=1}^N$. The training variables in the tensor-compressed model are the tensor cores $\{\ten{G}_i\}_{i=1}^N$. Suppose that the tensor-compressed model parameterized by the tensor cores is $f(\mat{x}|\{\ten{G}_i\}_{i=1}^N)$. The full-precision end-to-end tensor-compressed training is to minimize the model loss:
\vspace{-1mm}
$$
    \min_{\{\ten{G}_i\}_{i=1}^N} \sum_{k} \text{loss}\left(\text{target}_k, f(\mat{x}_k|\{\ten{G}_i\}_{i=1}^N)\right).
$$

\subsection{End-to-End Quantization-Aware Training} 
\label{subsec:quantize}
With TT/TTM compression, we further reduce the model size by quantization-aware training with learnable scale factors. The goal is to obtain ultra low-bit representation for all tensor cores used for compressing a transformer model.

Assume that the tensor cores are represented with $b$-bit quantization $\{Q(\ten{G}_i,\delta_i,b)\}_{i=1}^N$ to save the computing and memory cost on edge devices. The quantization-aware tensor-compressed training computes the tensor cores $\{\ten{G}_i\}_{i=1}^N$ and scales $\{\delta_i\}_{i=1}^N$ via solving the following optimization problem: 
\vspace{-1mm}
$$
    \min_{\{\ten{G}_i,\delta_i\}_{i=1}^N}\sum_{k} \text{loss}\left(\text{target}_k, f(\mat{x}_k|\{Q(\ten{G}_i,\delta_i,b) \}_{i=1}^N)\right).
$$
We observe that the tensor cores are well centered around $0$, thus a symmetric quantization with scaling is employed. The quantization function $Q$ is defined as 
\[
    Q(x,\delta,b) := \delta\, \text{round}\left(\text{clip}(\frac{x}{\delta},-2^{b-1},2^{b-1}-1)\right),
\]
where $\text{round}(a)$ rounds $a$ to its nearest integer and the function $\text{clip}(a,v_{\min},v_{\max})$ clips $x$ into the range $[v_{\min},v_{\max}]$. The quantization $Q(x,b,\delta)$ maps $x$ into $\texttt{INT}_b$. The range of weights and tensor cores may differ dramatically before and after training. Thus, we set the scaling factor $\delta$ as a learnable variable that can be automatically determined during training. The quantization function $Q$ is not differentiable, but we can compute the fake gradients to $\ten{G}_i$ and $\delta_i$ using straight-through estimators. According to \cite{jain2020trained}, the fake gradient of $Q(x,\delta,b)$ with respect to $\delta$ and $x$ is
\begin{equation}\label{eq:scale grad}
    \frac{\partial Q(x,\delta,b)}{\partial \delta}:=\left \{ 
    \begin{array}{ll}
       \frac{Q(x,\delta,b)-x}{\delta}  & \text{if } -2^{b-1}\le \frac{x}{\delta} \le 2^{b-1}-1 \\
        -2^{b-1} & \text{if } \frac{x}{\delta}<-2^{b-1} \\
        2^{b-1}-1 & \text{if } \frac{x}{\delta}>2^{b-1}-1
    \end{array}
    \right. 
\end{equation}
\begin{equation}
    \frac{\partial Q(x,\delta,b)}{\partial x}:=\left \{ \begin{array}{ll}
        1 & \text{if } -2^{b-1}\le \frac{x}{\delta} \le 2^{b-1}-1 \\
        0 & \text{otherwise}
    \end{array}
    \right. .
\end{equation}
The scaling factors are layer dependent, i.e., different linear layers have different scaling factors. All tensor cores in the same linear layer share the same scaling factor since they typically have a similar range. During quantization-aware training, the inputs of the linear layers are quantized into $\texttt{INT}_8$, and all intermediate tensor contractions are computed in $\texttt{INT}_8$ to reduce computation costs further.

\subsection{Layer-by-Layer Distillation of Low-Precision Tensor-Compressed Transformers}
\label{subsec:distillation}
Our quantization-aware tensor-compressed training performs very well on end-to-end tasks. However, in some practical NLU tasks, an end-to-end training from scratch can be expensive. Instead, one may want to learn a small-size model from a pre-trained large model. Therefore, this subsection presents a layer-by-layer distillation to learn a low-precision tensor-compressed student model from a fine-tuned teacher model. 

Suppose the teacher model is a transformer with an embedding and $L$ encoders. The low-precision tensor-compressed student model has the same model architecture as the teacher model, while the weights are represented by low-precision low-rank tensor cores. Let $\mat{y}_{\rm emb}$ be the output of the embedding table, $\mat{y}$ be the predicted soft label, and $\mat{y}_i,\mat{ attn}_i$ be the output and attention probability matrix of the $i$th encoder block of a transformer. The superscript $t$ and $s$ indicate a teacher model and a student model, respectively. Existing works \cite{jiao2019tinybert,zhang2020ternarybert,bai2020binarybert} use the following distillation loss for compression:
\begin{align}\label{eq:loss}
    \mathcal{L}_{\rm all}:=& {\rm MSE}(\mat{y}_{\rm emb}^t,\mat{y}_{\rm emb}^s) + {\rm COS}(\mat{y}_{\rm emb}^t,\mat{y}_{\rm emb}^s) \nonumber \\
         &  + \sum_{i=1}^L ({\rm MSE}(\mat{y}_i^t,\mat{y}_i^s)+{\rm COS}(\mat{y}_i^t,\mat{y}_i^s)) \nonumber\\
     &  + \sum_{i=1}^L {\rm CE}(\mat{ attn}_i^t,\mat{ attn}_i^s))  + {\rm CE}(\mat{y}^t/T,\mat{y}^s/T),
\end{align}
where ${\rm MSE}$ is the mean squared error, ${\rm COS}$ is the cosine similarity, ${\rm CE}$ is the cross entropy loss, and $T$ is the temperature of soft labels. The above distillation loss matches soft labels, the internal outputs, and attention probabilities to increase the generalization property of a student model. 

{\bf Address the Convergence Issue in Tensor-Compressed Settings.} Most existing works \cite{sanh2019distilbert,jiao2019tinybert,zhang2020ternarybert,bai2020binarybert} reuse pretrained weights in the teacher model to initialize the student model by making the two models similar at the beginning of training. Thus, the distillation loss containing outputs and attention probabilities of all layers performs well. However, the pretrained weight matrices may not have low-rank structures, thus they cannot be directly used to initialize tensor-compressed distillation. In fact, the initial tensor-compressed transformer is very different from the teacher model, causing the distillation loss of all layers to fail in tensor-compressed training. Motivated by this observation, we use the layer-by-layer distillation proposed in \cite{aguilar2020knowledge}, which matches the outputs and attention probabilities from top layers to bottom layers. The layer-by-layer distillation starts from the embedding table with loss $\mathcal{L}_0:={\rm MSE}(\mat{y}_{\rm emb}^t,\mat{y}_{\rm emb}^s) + {\rm COS}(\mat{y}_{\rm emb}^t,\mat{y}_{\rm emb}^s)$. Then, the loss for the $i$th encoder block is 
\begin{eqnarray*}
     \mathcal{L}_i&:=& \mathcal{L}_{i-1}+{\rm MSE}(\mat{y}_i^t,\mat{y}_i^s)+{\rm COS}(\mat{y}_i^t,\mat{y}_i^s)\\
     & & +  {\rm CE}(\mat{attn}_i^t,\mat{attn}_i^s)).
\end{eqnarray*}
The loss $\mathcal{L}_i$ aims to match the outputs and attention of the first $i$ encoder blocks. We train the tensor-compressed model using the losses $\mathcal{L}_0,\mathcal{L}_1,\ldots,\mathcal{L}_L$ sequentially. Finally, the soft labels are added to the loss $\mathcal{L}_L$, and the loss becomes $\mathcal{L}_{\rm all}$ in \eqref{eq:loss}.

\section{Experiments}
We use two natural language understanding (NLU) benchmarks to test our quantization-aware tensor-compressed training framework. Specifically, we test end-to-end and distillation-based training on the ATIS dataset \cite{hemphill1990atis} and the GLUE benchmark \cite{wang2018glue}, respectively. For the GLUE benchmark, we use the fine-tuned BERT \cite{devlin2019bert} model as a teacher model for our quantization-aware and tensor-compressed distillation.


\subsection{ATIS Dataset for End-to-End Training}
The airline travel information system (ATIS) dataset \cite{hemphill1990atis} is an NLU dataset containing utterances related to queries for flight reservations. For each utterance, we need to detect its intent and the slot annotation for each word in the utterance. On this dataset, we perform \textbf{end-to-end} quantization-aware tensor-compressed training. 

\begin{table}[t]
    \centering
    \footnotesize	
    \caption{Tensor-compression setting for ATIS dataset.}
    \label{tab:tensor ATIS}
    \begin{tabular}{|c|c|c|c|c|}
    \hline
     & format & linear shape &tensor shape & rank \\
     \hline
       embedding  & TTM & (800,768) & (15,20,16,16,8) & 30\\
        attention & TT & (768,768) & (24,32,32,24) & 10\\
        feed-forward & TT & (768,3072) & (32,24,48,64) & 10\\
        classification &TT & (768,768) & (24,32,32,24) &  10 \\
        \hline
    \end{tabular}
\end{table}

The transformer model for this task has one embedding table, two encoders, and two classification heads, where one head is for intent classification and the other one is for slot filling. 
We use its full-size and full-precision model as a baseline. 
We compress the embedding table and the two encoders into quantized tensor cores. The first linear layer of each classification is compressed into full-precision tensor cores. All other layers are kept in the original form. Table \ref{tab:tensor ATIS} lists the compressed tensor shapes and ranks. We use batch size $32$ and the Adam optimizer \cite{kingma2014adam} with $\beta_1=0.9,\beta_2=0.98$, and learning rate $10^{-3}$. For each model, we train 40 epochs and report the result in Table \ref{tab: ATIS result}.

The intent classification task is measured by accuracy, and the slot filling is measured by F1-score. The test results are reported in Table \ref{tab: ATIS result}. Our full-precision tensor-compressed model reaches $19\times$ size reduction with almost the same performance compared with the full-precision full-size baseline. The $\texttt{INT}_8$ and $\texttt{INT}_4$ models perform similarly to the baseline and the $\texttt{FP}_{32}$ tensor-compressed model, with less than 1\% accuracy and F1-score drop. The intent accuracy of the $\texttt{INT}_2$ model drops marginally. The $\texttt{INT}_4$ and $\texttt{INT}_2$ models have almost the same size, because the low-precision tensor cores consumes negligible memory and the uncompressed layers and parameters (e.g., layer normalization and bias vectors) dominate the storage cost. We can conclude that the quantized tensor-compressed transformer can reach around $60\times $compression ratio with less than 2\% accuracy drop on this dataset.

\begin{table}[t]
    \centering
    \caption{Tensor-compressed training of Transformer on ATIS dataset in precisions $\texttt{INT}_2$, $\texttt{INT}_4$, $\texttt{INT}_8$, and $\texttt{FP}_{32}$}
    \label{tab: ATIS result}
    \begin{tabular}{|c|c|c|c|}
    \hline
           & intent & slot & size (MB)\\
           \hline
          Full-size full-precision & 95.2  & 97.0  & 63 $(1\times)$ \\
          \hline
          Tensor-compressed $\texttt{FP}_{32}$ & 96.0  & 96.2  & 3.3 $(19\times)$ \\
          Tensor-compressed $\texttt{INT}_8$ & 95.5  & 96.1  & 1.4 $(45\times)$\\
         Tensor-compressed $\texttt{INT}_4$ & 94.3  & 96.2  & 1.1 $(57\times)$\\
         Tensor-compressed $\texttt{INT}_2$ & 93.6  & 95.0  & {\bf 1.0} $\mathbf{(63\times)}$\\
         \hline
      \end{tabular}
\end{table}

\subsection{GLUE Benchmark for Distillation}

The General Language Understanding Evaluation (GLUE) benchmark \cite{wang2018glue} is a collection of multiple natural language understanding tasks. It is widely used to evaluate the performance of natural language models. Four datasets in GLUE are chosen to test the proposed quantization-aware tensor-compressed {\bf distillation method} described in Section~\ref{subsec:distillation}. Among them, MNLI and QNLI have the largest size, SST-2 is moderate, and MRPC is the smallest. These datasets cover common natural language understanding tasks. 

 BERT-base is a large model containing one large embedding table and twelve encoders. One classification consisting of two linear layers is attached to the end of BERT. The embedding table, all linear layers in encoders, and the first linear layer in the classification are compressed via quantization-aware tensor-compressed training. We first fine-tune BERT-base on each dataset and use the fine-tuned BERTs as the teacher models for layer-by-layer distillation. Table \ref{tab:tensor GLUE} shows the detailed compression setting. In the experiments, we use batch size 32 and the Adam optimizer \cite{kingma2014adam} with $(\beta_1, \beta_2)=(0.9, 0.98)$. The learning rate is $10^{-3}$ for the losses $\mathcal{L}_0,\ldots,\mathcal{L}_{12}$ and is $5\times 10^{-5}$ for the last loss $\mathcal{L}_{\rm all}$. For each loss, we run 3, 5, 10, and 20 epochs for MNLI, QNLI, SST-2, and MRPC, respectively.

\begin{table}[b]
    \centering
    \footnotesize	
    \caption{Tensor-compression setting for BERT-base.}
    \label{tab:tensor GLUE}
    \begin{tabular}{|c|c|c|c|}
    \hline
     & format & linear shape &tensor shape  \\
     \hline
       embedding  & TTM & (30522,768) & (64,80,80,60) \\
        attention & TT & (768,768) & (24,32,32,24) \\
        feed-forward & TT & (768,3072) & (32,24,48,64) \\
        classification &TT & (768,768) & (24,32,32,24) \\
        \hline
    \end{tabular}
\end{table}

  \begin{table*}[t]
    \centering
    \caption{Distillation-based tensor-compressed training results on development split of the GLUE benchmark. The $\texttt{INT}_8$ tensor-compressed model has the same number of operations as $\texttt{FP}_{32}$, but those operations are cheap fixed-point operations. }
    \label{tab:result GLUE}
    \begin{tabular}{|c|c|c|c|c|c|c|c|c}
    \hline
    &precision & size (MB) & FLOPs (G) & MNLI & QNLI &SST-2  & MRPC \\
      \hline
    BERT-base \cite{devlin2019bert} & $\texttt{FP}_{32}$ & 423 (1$\times$) &20.3 ($1\times$) & 83.4 & 91.2 & 92.8 & 87.7 \\
    DistilBERT{\cite{sanh2019distilbert}} & $\texttt{FP}_{32}$ & 254 (1.7$\times$) &10.1 ($2\times$) & 82.2 & 89.2 & 91.3 & 87.5 \\
    BinaryBERT \cite{bai2020binarybert} & $\texttt{INT}_1$ & 16.5  (26$\times$) & 3.1 (7$\times$) & 84.2 & 91.5 & 92.6 & 85.5\\
    LadaBERT-4 \cite{mao-etal-2020-ladabert}& $\texttt{FP}_{32}$ & 42 (10 $\times $) & ---& 75.8 &75.1 & 84.0 & ---\\
        \hline
    \multirow{3}{*}{Rank 50}& $\texttt{FP}_{32}$ & 99 ($4\times$) &3.8 ($5\times$)  & 82.1 & 89.1 & 90.0 &  86.5 \\
    & $\texttt{INT}_8$ & 24.3 (17$\times$)&3.8 ($5\times$)& 80.7 & 88.1& 89.6& 85.8 \\ 
    & $\texttt{INT}_4$ & 12.1 (35$\times$)&1.9 ($11\times$)& 79.7 & 87.9 & 89.2  &85.5 \\
    \hline 
    \multirow{3}{*}{Rank 30} &$\texttt{FP}_{32}$ & 39 (11$\times$) &1.8 ($11\times$) & 80.1 & 88.1 & 89.3  & 85.1\\
     &$\texttt{INT}_8$ & 9.5 (45$\times$) & 1.8 ($11\times$)& 78.3 & 87.2 & 89.2  & 85.0\\
     &$\texttt{INT}_4$ & \textbf{4.8} ($\mathbf{88\times}$) & \textbf{0.9} ($\mathbf{23\times}$)& 77.4 & 86.9 & 88.3  &84.8 \\
     \hline

    \end{tabular}
  \end{table*}

Test results are reported in Table \ref{tab:result GLUE}. All tasks are measured by \textbf{accuracy}. We test two different ranks $30$ and $50$. The full precision tensor-compressed training of rank 50 maintains the most performance of BERT-base with only $1\%-2\%$ accuracy drop on every task. The accuracy slightly drops when decreasing the precision to $\texttt{INT}_8$ and $\texttt{INT}_4$ while the 
compression ratio increases to $17\times$ and $35\times$ from $4\times$. The $\texttt{INT}_4$ model is only 12.1MB, suitable for inference on middle resource-constrained edge devices. All results of rank 30 are slightly worse than rank 50 because of the smaller model size. The rank 30 model in $\texttt{INT}_4$ is only 4.8MB while still having acceptable accuracy. The tiny model is suitable for edge devices with strictly limited memory. The tensor-compressed training can easily adjust the model size by tuning the tensor rank in the model. It makes the quantized tensor-compressed transformer work for a wide range of devices with various resource budgets. In practice, we can also use rank-adaptive training~\cite{hawkins2022towards} to automatically determine the tensor ranks in both end-to-end training and distillation-based training.

The 4th column of Table \ref{tab:result GLUE} shows the estimated computational FLOPs at inference for each model. Here, we only count the FLOPs for matrix-vector/tensor-vector multiplications in encoders to simplify the computation. Other operations, like layer normalization and bias addition, only take a very small amount of computation compared to matrix-vector/tensor-vector multiplications. For $\texttt{FP}_{32}$ operations and quantized operations, FLOPs stand for the number of floating-point operations and fixed-point operations, respectively. We follow \cite{bai2020binarybert} to count the quantized operations, i.e., the multiplication between an $m$-bit number and an $n$-bit number roughly needs $\frac{mn}{64}$ fixed point operations. The full-precision tensor-compressed model saves $5\times$ and $11\times$ FLOPs for ranks $50$ and $30$, respectively. The $\texttt{INT}_8$ tensor-compressed model has the same number of operations as $\texttt{FP}_{32}$, but those operations are cheap fixed-point operations. After reducing the precision to $\texttt{INT}_4$, the saving of FLOPs further increases to $11\times$ and $23\times$ for ranks $50$ and $30$, respectively. 

{Compared to DistilBERT \cite{sanh2019distilbert}, BinaryBERT \cite{bai2020binarybert}, and LadaBERT \cite{mao-etal-2020-ladabert}, our quantization-aware tensor-compressed approach reaches the highest compression ratio ($88\times $) with little accuracy drop and has more flexibility to handle the trade-off between model performance and model size by tuning tensor ranks and model precisions.}


{We demonstrate the reduced computation of tensor-compressed training and inference by end-to-end training on the MNLI dataset training split with 393,000 sentences on an RTX-3090 GPU with 24G memory.} Table \ref{tab:time saving} shows that the tensor-compressed training and inference are $1.8\times$ faster than the uncompressed training. The time reduction ratio is less than the FLOPs reduction ratio in Table \ref{tab:result GLUE} because some small tensor contractions in tensor-vector multiplication are not parallelized on GPU. We expect the runtime reduction ratio to be similar to the FLOPs reduction ratio after optimizing the parallelization of the tensor contractions.




  \begin{table}
  \small
  \caption{Inference and training time for one epoch on MNLI training split with batch size 128.}
  \label{tab:time saving}
    \begin{tabular}{|c|c|c|}
    \hline
      & inference & training  \\
      \hline
    uncompressed& 8.8min & 26min  \\
    tensor-compressed& \textbf{4.7min} $(\mathbf{1.8\times})$ & \textbf{14min} 
    $\mathbf{(1.8\times)}$ \\
    \hline
    \end{tabular}
    
  \end{table}

\section{Conclusions and Remarks}
To compress transformer-based NLU models, we have proposed a quantization-aware and tensor-compressed method for both end-to-end training and distillation-based training. The embedding table and linear layers are compressed into small tensor cores, thereby substantially reducing the total number of model parameters. Besides that, we have applied quantization to each tensor core, further reducing memory costs. Quantization-aware training with trainable scaling factors has been used to learn the quantized tensor cores. To learn a compact NLU and speech recognition model from a pre-trained large transformer model, we have proposed to use layer-by-layer distillation method. This method outperforms the distillation that combines all layer outputs which typically leads to divergence in tensor-compressed training. We have evaluated our quantization-aware tensor-compressed training for two NLU tasks, where our compressed models have achieved high compression ratios with minimal accuracy drop. The quantized tensor-compressed models can have vastly different model sizes for various combinations of tensor ranks and precision. The experiment has demonstrated that our approach could maintain good accuracy even for extremely low ranks and precision. Our method allows additional deployment flexibility on devices with varying resource constraints. 

We would like to remark that our method can be applied to all transformer-based models for compression, not only limited to BERT. For instance, our approach has the potential to highly compress the transformer part of wav2vec2 \cite{baevski2020wav2vec}, a pre-trained transformer-based model for speech recognition. 

\section{Acknowledgements}

The authors would like to thank Ershad Banijamali, Clement Chung, Athanasios Mouchtaris, and Hieu Nguyen from Amazon for their fruitful suggestions and comments! 


\bibliographystyle{IEEEtran}
\bibliography{mybib}

\end{document}